# A Comparative Study of Convolutional and Recurrent Neural Networks for Storm Surge Prediction in Tampa Bay

1st Mandana Farhang Ghahfarokhi
*Department of Engineering Management and System Engineering*
*Old Dominion University*
Norfolk, USA
mfarh002@odu.edu

2nd Anahita Farhang Ghahfarokhi
*Faculty 2 of Computer Science and Engineering*
*Frankfurt University of Applied Sciences*
Frankfurt, Germany
a.farhangghahfa@alum.uca.es

3rd Mahta Zamanizadeh
*Department of Civil and Environmental Engineering*
*Old Dominion University*
Norfolk, USA
mzama003@odu.edu

4th Seyed Hossein Sonbolestan
*Department of Civil and Environmental Engineering*
*Old Dominion University*
Norfolk, USA
ssonb001@odu.edu

***Abstract*— In this paper, we compare the performance of three popular deep learning architectures: Long Short-Term Memory (LSTM), Convolutional Neural Networks (CNN), and CNN-LSTM in water level prediction over different time window sizes. We used wind, pressure, and tide data from National Oceanic and Atmospheric Administration (NOAA) stations. Our results show that the LSTM model generally outperformed the other models in forecasting water level trends. By applying LSTM, we achieved a mean squared error (MSE) of 0.006, making it the most reliable option for real-world use. We predicted water levels using wind, pressure, and tides. Wind, pressure, and tide data are time-based features that lack spatial patterns. Consequently, LSTM performed better than spatial methods, such as CNN. We also investigated Hurricane Ian, which occurred in Tampa Bay in September 2022 and caused a significant storm surge. Our findings show that CNN and CNN-LSTM performed slightly better than LSTMs during powerful storms, such as hurricanes. Hurricanes are seldom observed in training data, and CNN-based models are more reliable when predicting data that were not included in the training. To explain how environmental factors affect the predicted water level, we used the perturbation method. The perturbation shows that the tide is more effective at predicting water levels than wind and pressure. This study demonstrates the potential of deep learning to improve storm surge prediction accuracy.**

## I. Introduction

The rise in flood heights is driven by climate change, and coastal regions, such as the Gulf of Mexico, are particularly affected. As global temperatures increase, sea levels also increase, leading to increased flooding from storm surges [1]. Storm surges, which are the abnormal rise in seawater levels due to storms, pose risks to human life and infrastructure [2]. Higher and more destructive storm surges are a direct consequence of the increasing intensity of hurricanes [3]. This growing threat highlights the need for accurate predictive models to mitigate the impact of these natural disasters on communities along the Gulf Coast [4].

Deep learning has proven to be a powerful method for predicting storm surges [5]. Recent studies have shown that models, such as Convolutional Neural Networks (CNNs) and Long Short-Term Memory (LSTM) networks can capture complex relationships in meteorological and oceanographic data with high accuracy [6]. By learning patterns in wind speed, atmospheric pressure, and previous sea level, these models predict the magnitude and timing of storm surges with notable accuracy [7]. In [6], the authors built a deep-learning surrogate model for coastal water levels (relevant to storm surges) to replace expensive numerical models. Spatiotemporal neural networks are used to predict water levels faster than traditional hydrodynamic models. In [8], the authors presented a causal spatiotemporal graph neural network that uses both station-to-station connections and surge dynamics. This model forecasts storm surges at multiple locations when there is limited observational data. By embedding causal relationships directly into the graph structure, their approach remains reliable even with sparse, noisy water level records. In [9], the authors used deep neural networks to predict short-term storm-surge levels at estuarine and coastal stations in the Pearl River region. This study showed that data-driven surge models provide reliable predictions in different locations regarding lead times. Using deep learning for storm surge prediction has advantages, such as faster computation compared to traditional numerical models [10]. This approach provides emergency response teams and coastal planners with a useful tool. It supports resource allocation and speeds up response times. This reduces the risk to human life and property.

The vulnerability of the Gulf of Mexico to natural disasters is increasing [11]. In this study, we selected a single tide gauge station in Tampa Bay, located in the Gulf of Mexico. Tampa Bay often experiences storm surges [12], making it an ideal use case to assess the effectiveness of surge prediction models. Although this study focuses on a specific region, it demonstrates the application of deep learning models to surge prediction and serves as a starting point for future multi-station studies. Previous studies have also examined sea levels in Tampa Bay [12, 13, 14]. In [13], the authors created a Tampa Bay Coastal Ocean Model (TBCOM). This is a physics-based system for nowcasting and forecasting that relies on the Finite Volume Community Ocean Model (FVCOM). The model incorporates wind, sea level, velocity, temperature, salinity, tides, and river discharge, and simulates the circulation of the coastal ocean to predict sea level and currents in Tampa Bay. In [12], the authors used the FVCOM to study hurricane storm surge in the Tampa Bay region and investigate the sensitivity of the storm surge to the point of landfall, direction, approach speed, and intensity. In [14], the authors simulated flood risk by identifying critical nonlinear relationships between flood-damage locations and flood-risk factors. Five machine learning models, including decision tree, support vector machine, adaptive boosting, extreme gradient boosting, and random forest, were used. They used 16 flood-risk factors, including distance to water bodies and rainfall.

In this study, we compared the following architectures for water level prediction: CNN, LSTM, and CNN-LSTM. These architectures were evaluated based on their prediction accuracy. Wind, pressure, and tidal information were used as the input data in our experiments. Furthermore, we used the perturbation-based Explainable Artificial Intelligence (XAI) method to determine the contribution of each feature (wind, pressure, and tide) to water level prediction.

The remainder of this paper is organized as follows: Section II provides details about our case study, that is, Tampa Bay. Section III presents our methodology, which uses three deep learning architectures. Section IV presents the experiments and results. Finally, Section V concludes the study and provides directions for future work.

## II. Case Study

In this study, we used the following data from Tampa Bay[1], as shown in Fig. 1, as inputs for our model:

- Wind: The wind data consist of two components:
    I. U component: The east-west component of the wind. If the wind blows from west to east, this will be positive, and in the opposite direction, it will be negative.[2]
    II. V component: The north-south component of the wind. If the wind blows from south to north, this will be positive, and in the opposite direction, it will be negative.
- Pressure: This is often surface pressure.
- Tide: We used tide height information extracted from recorded data from station 8726520 (St. Petersburg, FL, USA) on the western side of Tampa Bay. Petersburg, FL) on the western side of Tampa Bay.[3]

Wind and pressure data were extracted from a box corresponding to a region with a size of 3° × 3°, located near Tampa Bay (longitude (x-axis): −85° to −82° and latitude (y-axis): 26° to 29°). Given the spatial resolution of the data used, a region with a size of 3° × 3° results in a grid of approximately 15 × 15 points. Using wind, pressure, and tide data, we predict verified water level measurements at station 8726520 (St. Petersburg, FL) on the west side of Tampa Bay.

## III. Method

In physics-based models, fluid flow is governed by the conservation of mass and momentum, which are commonly formulated using the Navier–Stokes equations [15]. Modelling circulation and shallow domains often involves shallow water equations, which are variations of the Navier–Stokes equations [16]. For storm surge modelling, these equations are particularly effective because they simplify the Navier–Stokes equations by depth-averaging and assuming hydrostatic pressure [17]. Decades of research have presented various numerical methods for solving these equations [18]. The shallow water equations can be expressed as:

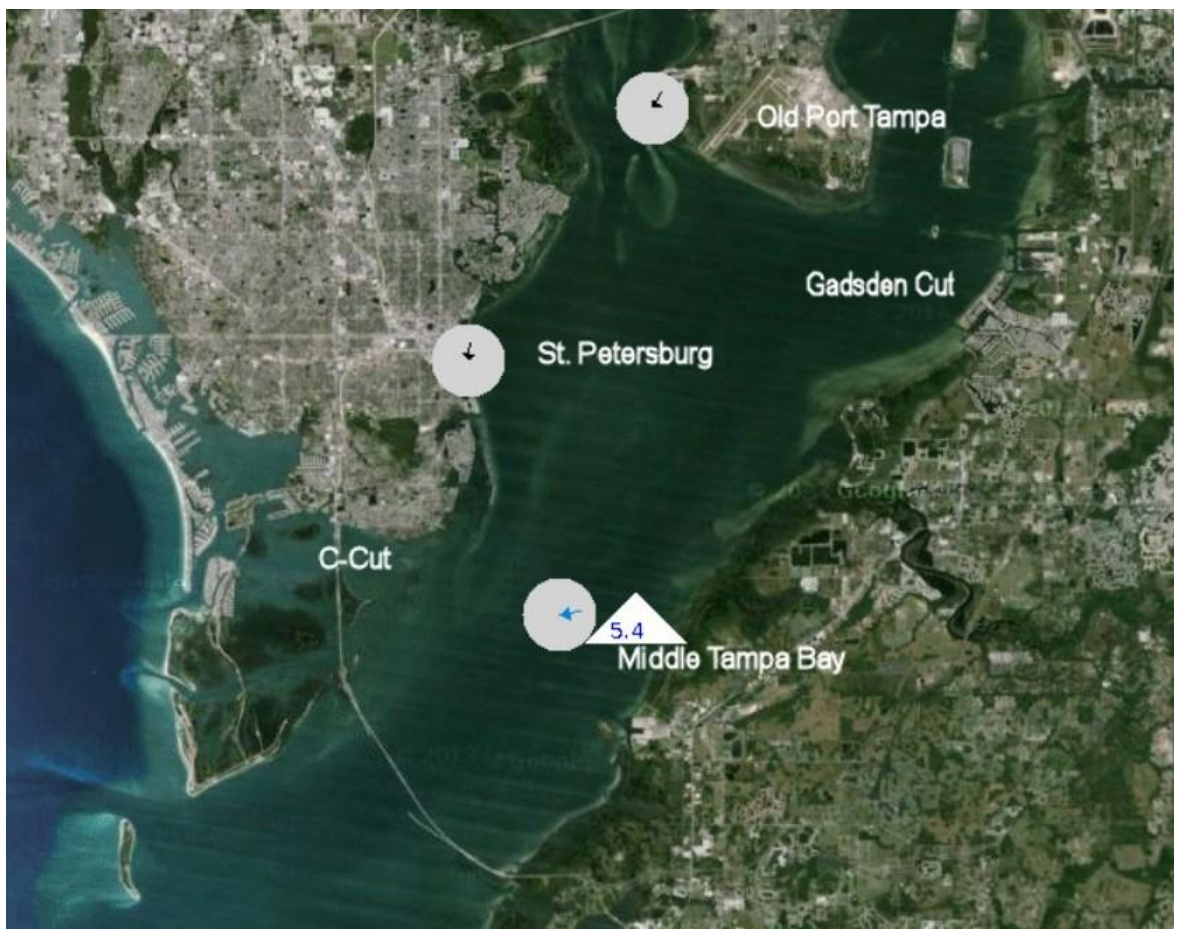

Figure 1: Tampa Bay. We have used information from station 8726520 (St. Petersburg, FL) on the Western side of Tampa Bay.

$$\frac{\partial U}{\partial t} + \frac{\partial F(U)}{\partial x} + \frac{\partial G(U)}{\partial y} = S(U) \quad (1)$$

Where $U$ is a matrix representing continuity and two horizontal momenta $[h, hu, hv]$, where $h$ is the depth of the water and $u$ and $v$ are the components of the horizontal velocity. $F(U)$ and $G(U)$ are the fluxes in the $x$ and $y$ directions, respectively, which include terms for linear and nonlinear advection, also known as Burgers' equation, pressure gradient and viscous terms. $S(U)$ represents the source term that accounts for external forces.

These equations are presented to show the complexity of the problem from a physics perspective. Moreover, they are computationally expensive to solve, mainly because of the complexity involved in solving the pressure equation, which in the case of shallow water reduces to the water level [15]. Additionally, improving accuracy often requires solving the advection terms using higher-order methods [19], thereby increasing computational costs. These factors make physics-based methods relatively slow. Machine-learning approaches are useful in this regard, as they offer faster predictions.

### A. *Machine Learning-Based Approach*

Machine learning (ML) models approach the problem differently by fitting an approximation function [20]. The parameters or weights of this function are determined during training. The optimization algorithm minimizes the difference between the predicted and true values by adjusting the weights [21]. Although ML models lack the interpretability of physics-based models, they offer major benefits in terms of computational efficiency [22]. Mathematically, an ML model can be viewed as a sequence of functions applied to the input data, which can be expressed as [23]:

$$\hat{y} = f_L(f_{L-1}(\ldots f_2(f_1(x;\ \theta_1);\ \theta_2)\ \ldots;\ \theta_{L-1});\ \theta_L) \quad (2)$$

[1] https://tidesandcurrents.noaa.gov/ports

[2] Wind and pressure data are collected from https://data.ucar.edu/dataset/ncep-climate-forecast-system-version-2-cfsv2-selected-hourly-time-series-products

[3] Tide and water level data are collected from https://api.tidesandcurrents.noaa.gov/api/prod/

Where $x$ denotes the input data, $\theta_i$ denotes the weights of layer $i$, $f_i$ denotes the activation function of layer $i$, and $\hat{y}$ denotes the predicted output. Training a deep learning model involves minimizing a loss function that measures the difference between the predicted and true values. In this study, we used the Mean Squared Error (MSE) as the loss function.

In addition to traditional machine learning, deep learning architectures offer stronger models for capturing temporal and spatial representations [24]. In this study, we employed three deep learning architectures CNN-LSTM, 3D Convolutional Neural Network (3D-CNN), and LSTM on the same training data and compared their performances.

### B. *Architectures of the Compared Models*

The atmospheric input to the model has dimensions of (36, 15, 15, 3). Here, 36 represents the size of the time window. This is optional and can be changed during the experiments. The dimensions of the 15 × 15 grid cover the geographical area discussed in Section II. The 3 indicates the components u and v of the wind data and the surface air pressure p, as discussed in Section II. The model's output is a time series of water levels at a single station in St. Petersburg, showing a water level of (36, 1). Additionally, the tidal input is fused into the model after the initial processing of the wind data.

#### *1) LSTM Architecture:*

Fig. 2 illustrates the LSTM-based architecture used in this study. The atmospheric input has dimensions (36, 15, 15, 3), which are reshaped at each time step into vectors of length 675 to form a shape input sequence (36, 675). This enables the direct processing of atmospheric data using recurrent layers. The reshaped sequence passes through an LSTM layer with 128 units, followed by a dense layer with 16 units and dropout, and then a second LSTM layer with 128 units. A parallel input branch provides the tidal signal with a shape of (36, 1), which is concatenated with the atmospheric representation. After concatenation, a dense layer with 16 units and dropout further refined the combined features, and a final linear dense layer outputted the predicted water level sequence with a shape of (36, 1). This architecture jointly incorporates atmospheric and tidal information for short-term water level forecasting.

#### *2) 3D-CNN Architecture*

Fig. 3 illustrates the 3D-CNN architecture. The atmospheric input is first passed through three Conv3D layers with 64, 32, and 16 channels, respectively. Each layer is followed by batch normalization and ReLU activation. Subsequently, the extracted features are reshaped and fed into two dense layers with 64 and 32 units, respectively. This processed atmospheric representation is then concatenated with the tidal input from a separate branch. The combined features pass through a 16-unit density layer and a final linear output layer to produce timestep-wise predictions. This architecture evaluates whether 3D convolutions, which capture both spatial and temporal patterns, can outperform LSTM-based models.

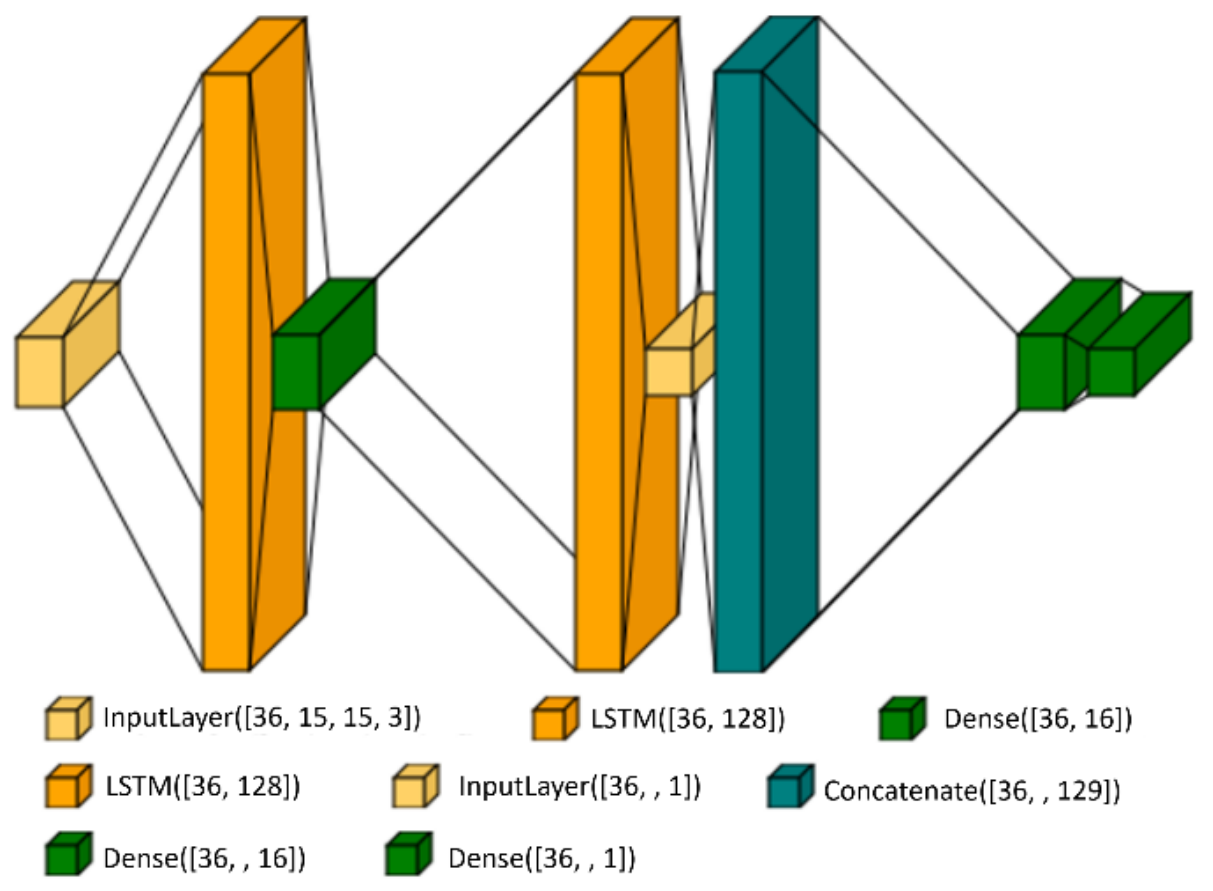

Figure 2: Abstract of the LSTM architecture that utilizes LSTM layers to process both atmospheric and tidal data.

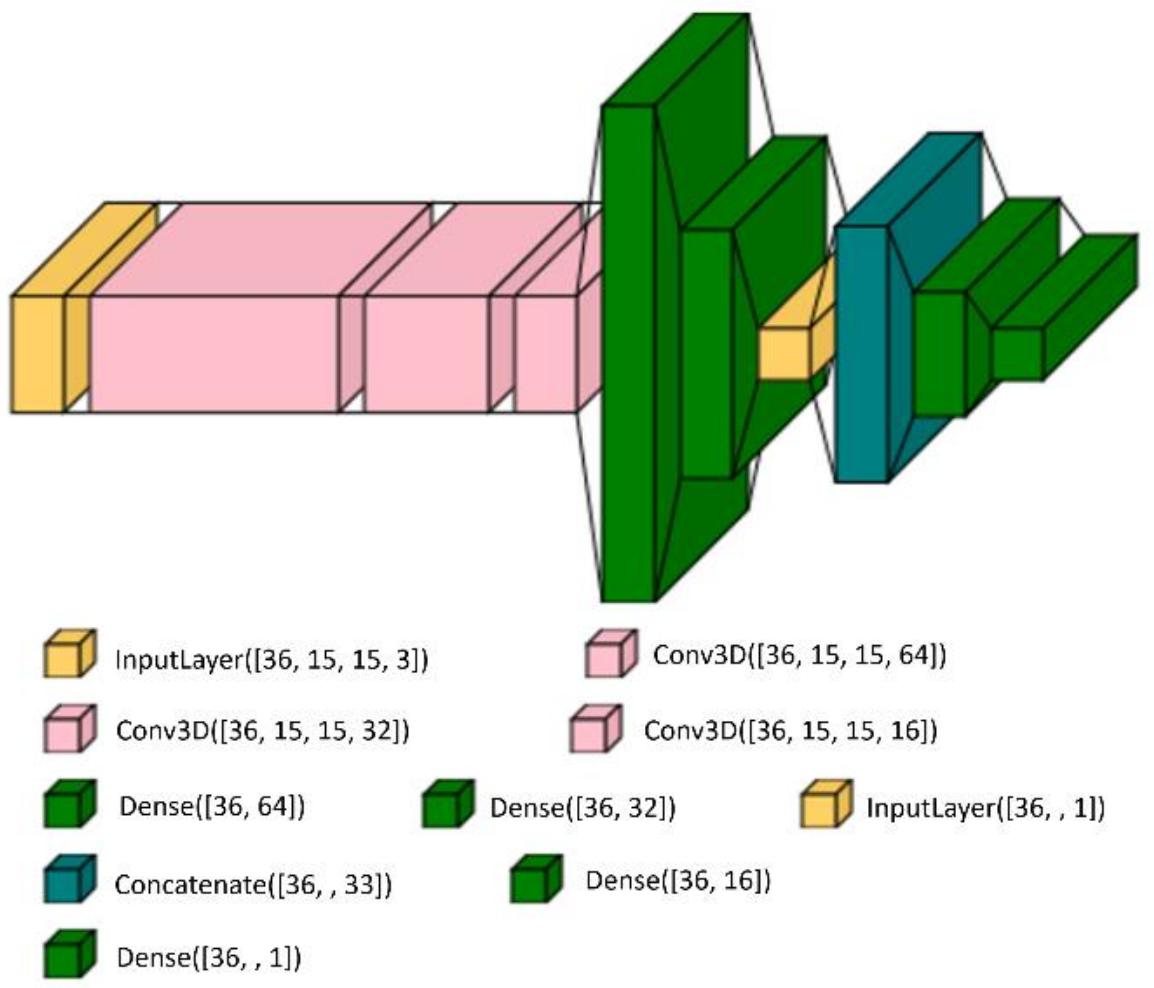

Figure 3: Abstract of the 3D-CNN architecture that employs 3D convolutional layers to capture spatial features directly.

#### *3) CNN-LSTM Architecture*

Fig. 4 shows the CNN– LSTM architecture used herein. The input is a tensor of size (36, 15, 15, 3). We extract spatial features using two time-distributed Conv2D layers with 64 and 32 filters, each followed by batch normalization and ReLU activation. The CNN output is reshaped and compressed through two dense layers with 128 and 32 units, including a dropout layer for regularization. Thereafter, an LSTM layer with 256 units is used to capture temporal dependencies. Its output is concatenated with a tidal input of shape (36, 1) and passed through a dense layer with 32 units before a final linear layer produces a prediction of size (36, 1), representing the time series of the water level.

## IV. EXPERIMENTS

Data were recorded hourly from 2011. All input variables (wind components, pressure, and tide heights) and the target variable (verified water level) were normalized by min-max scaling in the range [0, 1]. We used data before 2020 for training, the first half of the data after 2020 for validation, and the second half of the data after 2020 for testing.

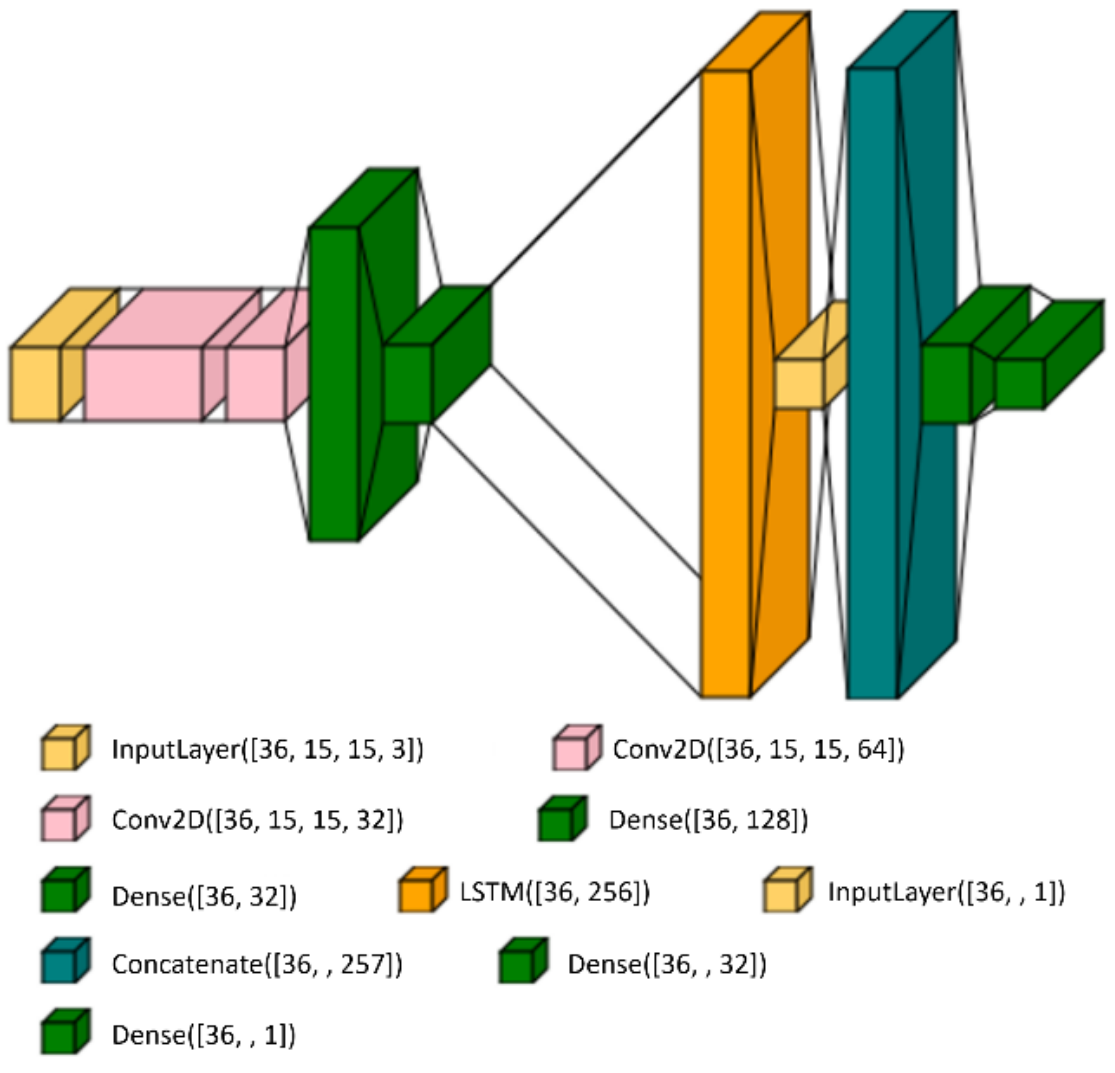

Figure 4: Abstract of the CNN-LSTM architecture. We used a combination of CNN and LSTM layers. CNNs capture spatial features, and LSTM layers focus on sequential dependencies.

The experiments were run on Google Colab using an A100GPU and the code is publicity accessible at https://github.com/AnaFarhang/A-Comparative-Study-of-CNNs-and-LSTMs-for-Storm-Surge-Prediction. First, we discuss the prediction results and present the MSEs for the general water trend and hurricane predictions. Next, we explain the effectiveness of each feature in the prediction.

## *A. Prediction*

We have applied three models discussed in Section III for different time windows ranging from 12 to 96 hours. To address the stochastic nature of neural networks and variations in results, we used a holdout method. For each model architecture, training was repeated five times, and we recorded the final values by averaging the results across these runs. The average MSE of the five runs of the program is shown in Table I and Fig. 5. In each run, we considered epochs = 50. As shown, with a window size of 24, we obtained the lowest MSE across all three methods. Moreover, window sizes that cover the entire day, such as 24, 48, 72, and 96, yield better results than the corresponding window sizes 12, 36, 60, and 84. This may be because whole-day data capture wind, pressure, or tide oscillations more completely than half-day data. The best result was obtained with a window size of 24. Therefore, we show the training and validation losses for a window size of 24 in Figs. 6 and 7. In general, we compared the methods and calculated average values across all window sizes in Table I, which shows that LSTM performs better than the other methods. The data used in this study (wind, pressure, and tide) are time series data, and because they represent temporal sequences, the LSTM architecture is particularly well-suited for modelling such time-dependent patterns. Unlike CNNs, which primarily capture spatial dependencies through convolutional filters, LSTMs are designed to capture long-term temporal dependencies using memory cells and recurrent structures.

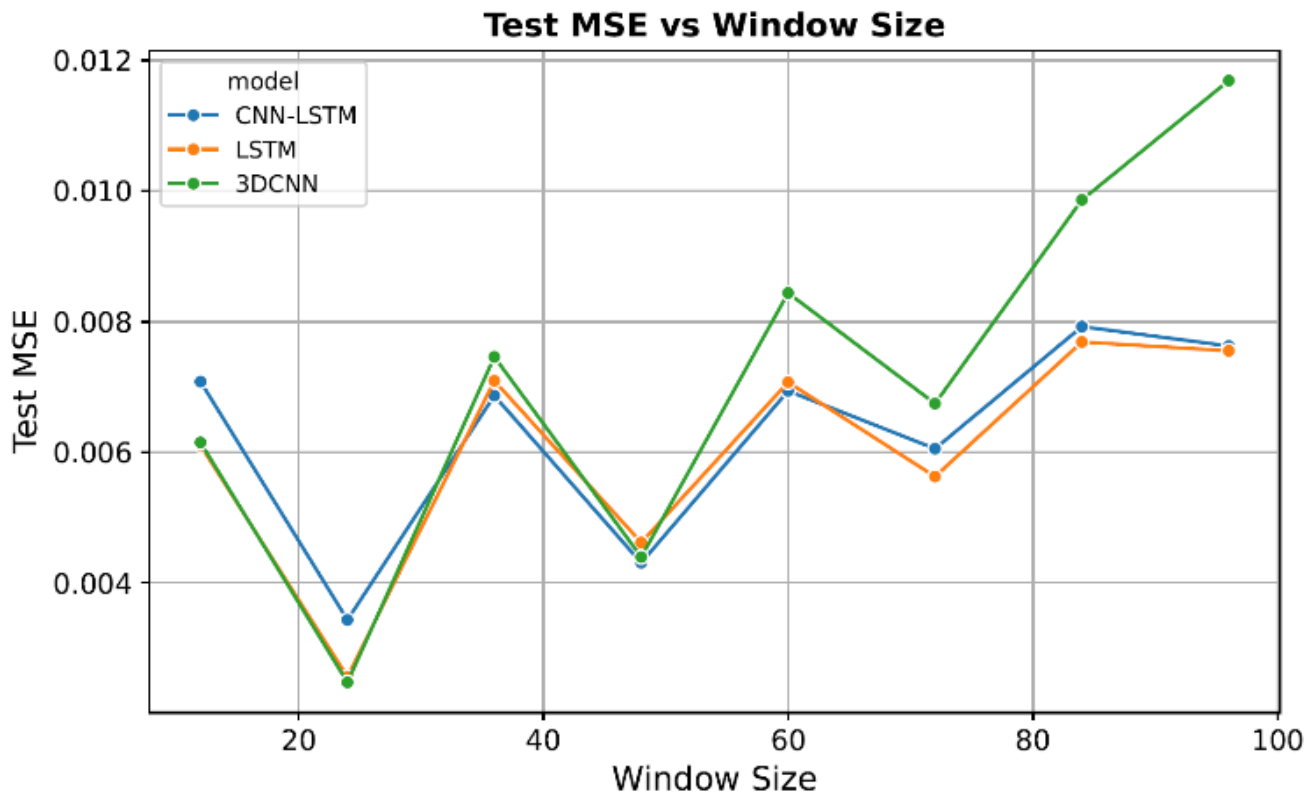

Figure 5: Test MSE versus window size. The x and y axes represent the window size used for prediction and the test MSE, respectively. We observe a general increase in the MSE as the window size increases.

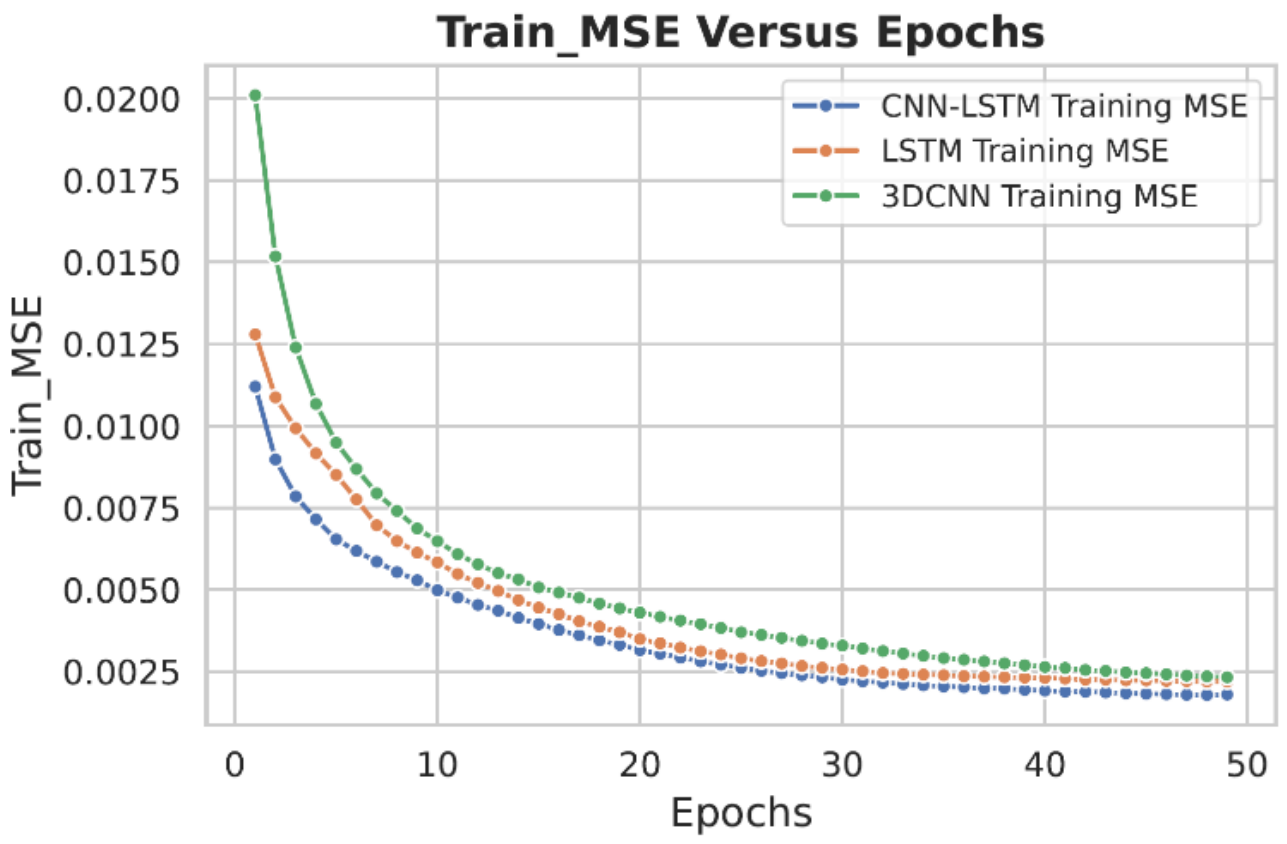

Figure 6: Training MSE versus epochs. We observe a decrease in the training MSE as the number of epochs increases.

We conducted a special study of Hurricane Ian, which occurred in September 2022 in Florida, and evaluated the proposed methods. We considered the start time window for the hurricane as 2022-09-27T12:00:00 and the end time as 2022-09-30T00:00:00 (i.e., the prediction extends 24 hours beyond 30$^{th}$ September). The predicted water level is shown in Fig. 8 and the MSE values are depicted in Table II. As we can see, there is an increase in MSE for all models compared to the values recorded in Table I. As shown in Table II, CNN and CNN-LSTM perform better than the LSTM model. Predicting hurricanes is tough for machine learning models because they often haven't seen hurricane examples in their training data to learn their patterns.

## *B. Explanation: Feature Importance*

For perturbation, we focused on a window size of 24 because it had the lowest average MSE among the others. We omitted each feature and built the model using the remaining features. The MSE test data were then recomputed. Table III lists the MSE values of the test data after removing each feature per model. The tide height was the most effective feature across all models. Fig. 9 shows the average MSE for all models after perturbation, revealing the effect of tide height on the prediction.

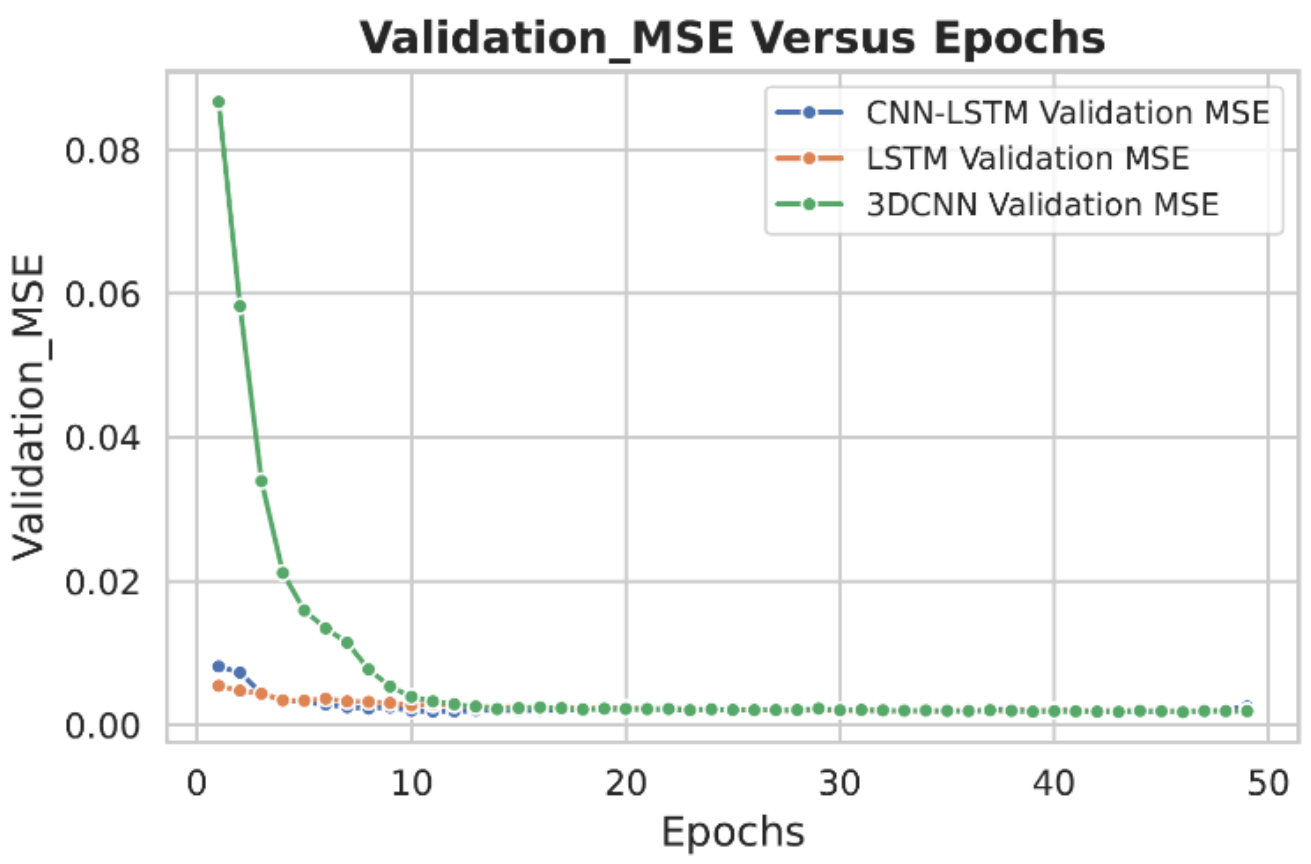

Figure 7: MSE of the validation set. We observed a decrease in the validation MSE as the number of epochs increased.

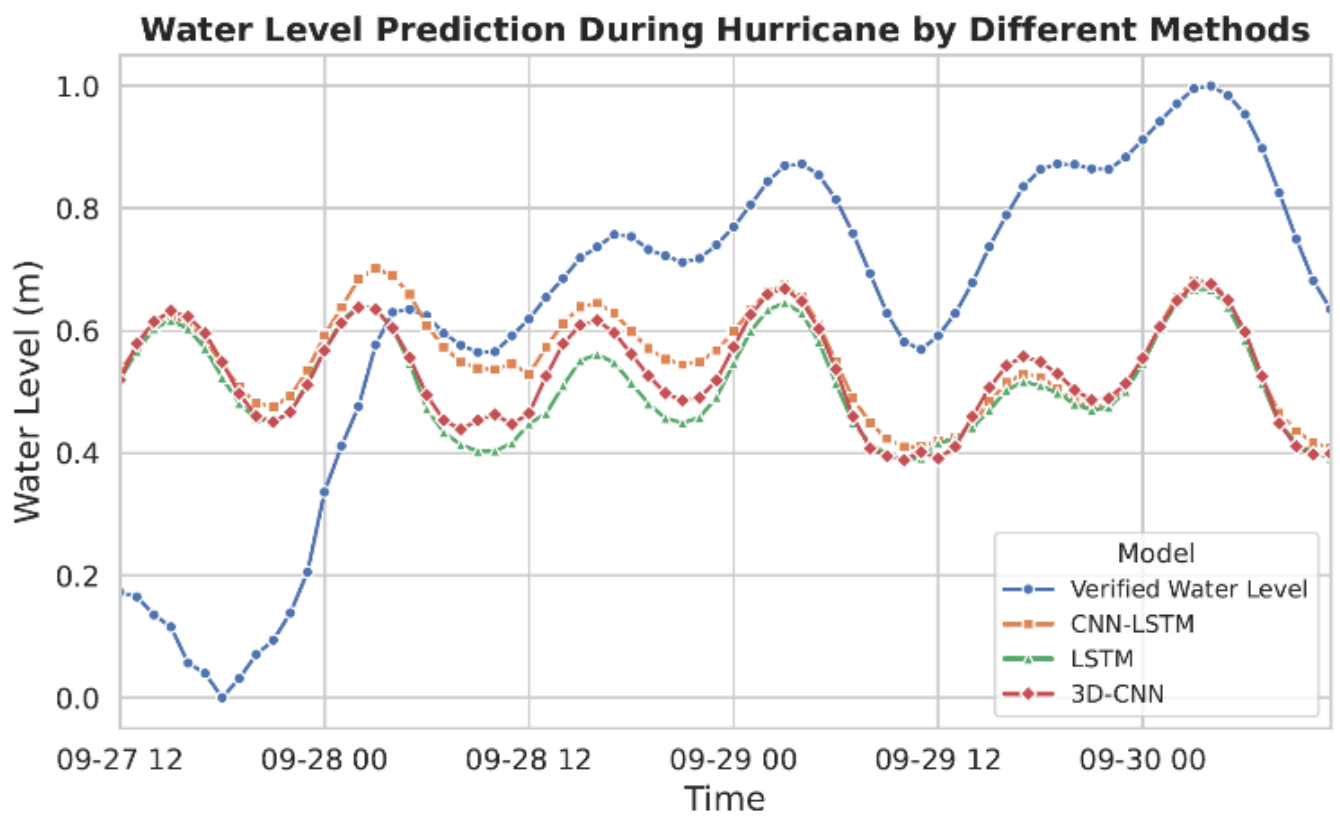

Figure 8: Predicted water level in Hurricane Ian. Here, we present the predicted water levels from three models and the actual water levels during the hurricane.

Table 1: Test MSE versus window size with average per model and window size. The MSE for each method across all time windows is shown.

| Model\Window size | 12 | 24 | 36 | 48 | 60 | 72 | 84 | 96 | Average |
|---|---|---|---|---|---|---|---|---|---|
| CNN-LSTM | 0.0071 | 0.0034 | 0.0069 | 0.0043 | 0.0069 | 0.0061 | 0.0079 | 0.0076 | 0.0063 |
| LSTM | 0.0061 | 0.0026 | 0.0071 | 0.0046 | 0.0071 | 0.0056 | 0.0077 | 0.0076 | 0.006 |
| 3DCNN | 0.0061 | 0.0025 | 0.0075 | 0.0044 | 0.0084 | 0.0067 | 0.0099 | 0.0117 | 0.0072 |
| Average | 0.0064 | 0.0028 | 0.0071 | 0.0044 | 0.0075 | 0.0061 | 0.0085 | 0.009 | 0.0065 |

Table 2: MSE values for Hurricane Ian, which occurred in Florida in September. We used a model that was previously trained for the time when the hurricane occurred and recorded the MSE values.

| Model | Hurricane MSE |
|---|---|
| CNN-LSTM | 0.081500 |
| LSTM | 0.090300 |
| 3DCNN | 0.084500 |

Table 3: Test MSE after perturbation. We removed each feature and built a model using the remaining features. Here, we show the test MSEs of the three models and their averages after removing each feature.

| Model | Pressure | Wind (U) | Wind (V) | Tide height |
|---|---|---|---|---|
| CNN-LSTM | 0.0028 | 0.0024 | 0.0027 | 0.0076 |
| LSTM | 0.0026 | 0.0028 | 0.0026 | 0.0061 |
| 3DCNN | 0.0026 | 0.0027 | 0.0025 | 0.0066 |
| Average | 0.0027 | 0.0026 | 0.0026 | 0.0068 |

## V. Conclusion

This study presents a comparative analysis of three deep learning architectures–CNN-LSTM, LSTM, and 3D-CNN–for the prediction of water levels with different window sizes. We used Tampa Bay wind, pressure, and tide data to predict water levels using the aforementioned models. Our findings show that all three models performed closely, with the best result achieved with a window size of 24 h. The LSTM performed slightly better at predicting water levels under normal conditions. Because time-based data, such as wind and pressure, were used as inputs, and they lacked spatial patterns, which are more suitable for CNNs. CNN and CNN-LSTM models performed slightly better under hurricane conditions. Because CNN-based models perform better with unseen data, and hurricanes are a type of data that is rarely seen in the training data. Moreover, we employed the perturbation method to identify the most effective features for prediction.

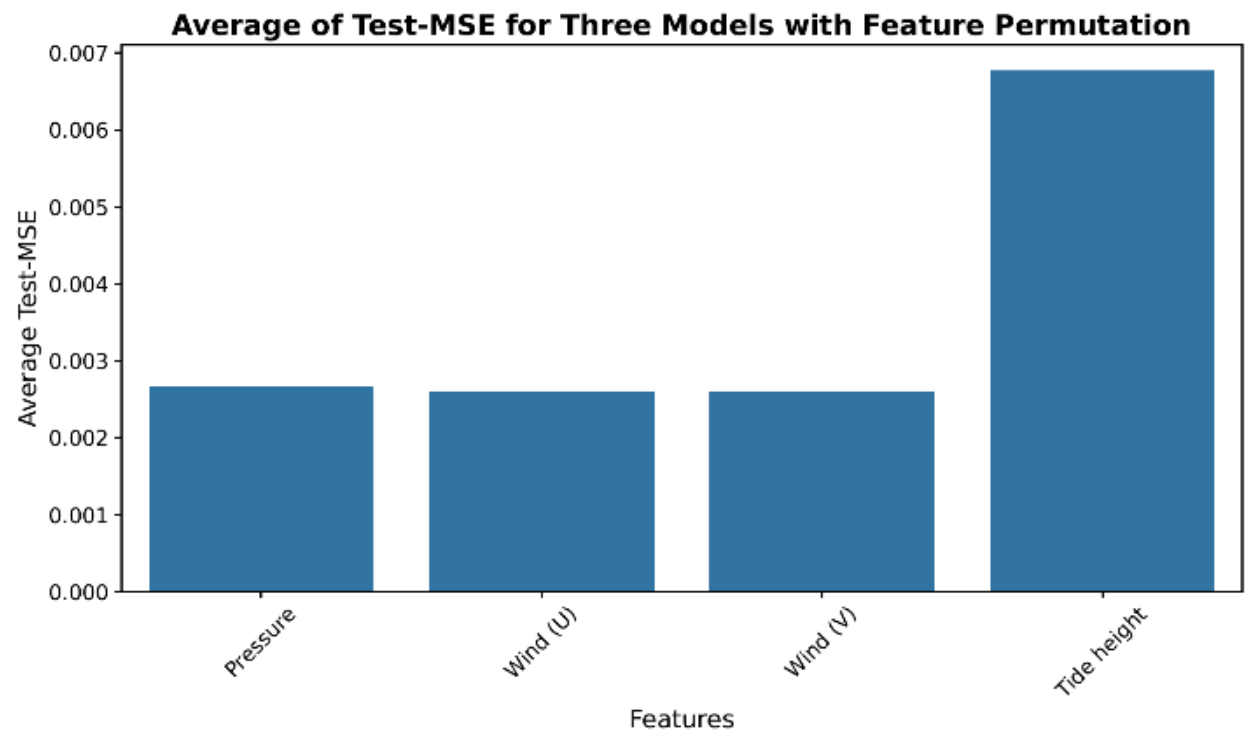

Figure 9: Average test MSE after perturbation. We removed the specified feature and built the model using the remaining features. Here, we present the average test MSEs of the three models after removing each feature.

The XAI results show that tide height is the most important feature for predicting water levels, followed by wind components and pressure. The insights gained from this study can contribute to the development of forecasting tools, helping emergency planners and policymakers respond quickly in the event of danger to human life and property.

## VI. FUTURE WORK

In this study, we considered data from a single tide station. However, this study can be extended to include information from multiple stations. This will improve the applicability of storm surge prediction models for coastal regions with multiple tidal stations. In addition, we applied deep learning methods to predict water levels. However, as presented in this study, physical-based equations (e.g., the Navier–Stokes equations) can also be used to predict water levels. It would be interesting to embed physical-based equations in deep learning methods to predict water levels. Beyond methodological improvements, water level prediction models can be used in critical decision-making systems. Therefore, potential threats must be considered. For example, ensuring the consistency of input data (wind, pressure, and tide measurements) is necessary). In these types of predictions, adversarial attacks such as data poisoning and sensor failure may occur. The XAI method that we considered helps mitigate such threats. It is also necessary to consider other types of threats, such as model envision attacks and communication attacks in sensor networks, and to understand their effects on the prediction. Finally, more advanced time series-based models, such as transformers, can be employed. Deep learning models used for water level prediction require scalable processing. Using edges and clouds together helps the system to work faster and reduces delays [25].